%% file: 0_main.tex
\pdfoutput=1

\documentclass[11pt]{article}

\usepackage[]{acl}

\usepackage{times}
\usepackage{latexsym}

\usepackage[T1]{fontenc}

\usepackage[utf8]{inputenc}

\usepackage{microtype}

\usepackage{inconsolata}

\usepackage{graphicx}

\usepackage{bm}

\usepackage{booktabs}
\usepackage{enumitem}
\usepackage{fancybox}

\newcommand{\sixsix}{\ensuremath{6\! \times\! 6}}

\title{Explaining Puzzle Solutions in Natural Language:\\ An Exploratory Study on $\bm{\sixsix}${} Sudoku}

\author{Anirudh Maiya \;\; 
Razan Alghamdi \\
\textbf{Maria Leonor Pacheco} \;\; 
\textbf{Ashutosh Trivedi} \;\; 
\textbf{Fabio Somenzi}  \\ 
University of Colorado Boulder \\
\texttt{\{first.last\}@colorado.edu} }

\begin{document}
\maketitle
\begin{abstract}
The success of Large Language Models (LLMs) in human-AI collaborative decision-making hinges on their ability to provide trustworthy, gradual, and tailored explanations.
Solving complex puzzles, such as Sudoku, offers a canonical example of this collaboration, where clear and customized explanations often hold greater importance than the final solution.
In this study, we evaluate the performance of five LLMs in solving and explaining \sixsix{} Sudoku puzzles. While one LLM demonstrates limited success in solving puzzles, none can explain the solution process in a manner that reflects strategic reasoning or intuitive problem-solving.
These findings underscore significant challenges that must be addressed before LLMs can become effective partners in human-AI collaborative decision-making.
\end{abstract}

\input{1_intro}
\input{2_related}
\input{3_task}
\input{4_experiments}
\input{5_conclusion}
\input{limitations}
\input{acknowledgments}

\bibliography{custom}

\appendix

\section{Puzzle difficulty rubric}\label{app:diff_rubric}
\begin{enumerate}
    \item \underline{Easy} (ER 1.0-1.2): Requires only hidden singles in blocks, making up 73.4\% of the dataset. These puzzles can be solved using basic scanning techniques. 
    \item \underline{Medium} (ER 1.5): Requires hidden singles in rows/columns, comprising 20.1\% of puzzles. These puzzles need slightly more advanced pattern recognition.
    \item \underline{Hard} (ER 1.7-2.5): No candidates needed but requires more complex deduction, representing 1.6\% of the dataset.
    \item  \underline{Fiendish} (ER 2.6-6.0): Requires writing down candidates and advanced techniques, making up 3.4\% of puzzles. 
    \item \underline{Diabolical} (ER 6.2+): Requires forcing chains and sophisticated solving strategies, comprising 1.5\% of the dataset.
\end{enumerate}

\begin{figure}[t]
    \centering
    \includegraphics[width=\linewidth]{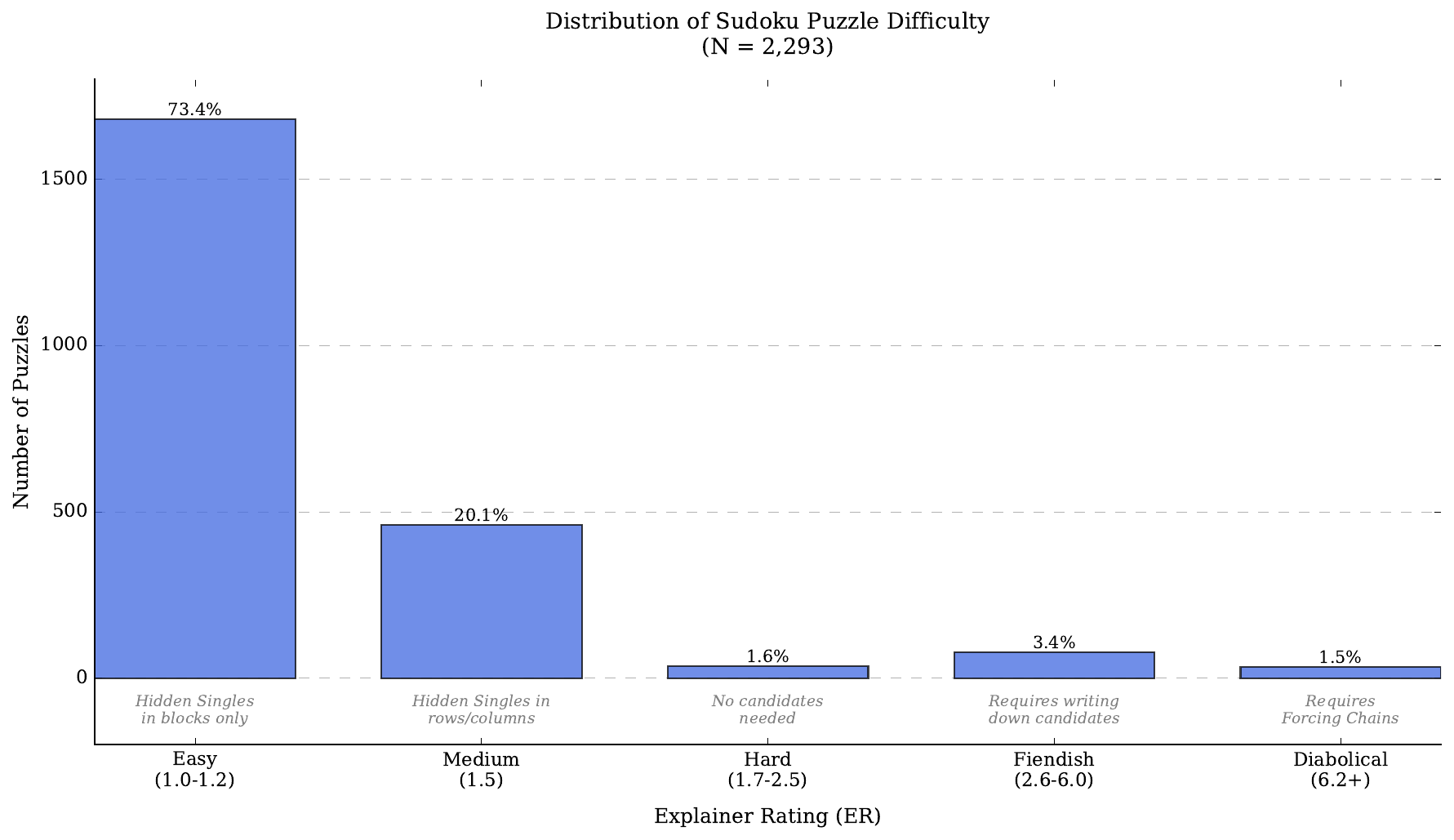}
    \caption{Distribution of Sudoku puzzle difficulty across the dataset (N = 2,293). Each category corresponds to specific solving techniques required, from basic hidden singles to complex forcing chains.}
    \label{fig:difficulty-dist}
\end{figure} 

\section{Prompt Details}
\label{sec:promptdetails}
\noindent\makebox[\linewidth]{%
    \ovalbox{
        \begin{minipage}{0.95\columnwidth}
\textbf{Task Description:} You are given an incomplete \sixsix{} Sudoku grid. The objective is to fill in the empty cells (represented by 0s) with numbers from 1 to 6 while following these rules:

\textbf{Row Constraint:} Each number from 1 to 6 must appear exactly once in every row.

\textbf{Column Constraint:} Each number from 1 to 6 must appear exactly once in every column.

\textbf{Box Constraint:} Each number from 1 to 6 must appear exactly once in each of the six $2\! \times\! 3$ boxes.

\textbf{Important:} Do not use any code to solve this puzzle. Solve this puzzle using logical reasoning and only provide the final grid.

\textbf{Input Format:}
A \sixsix{} grid is represented by six lines of input.
Each line contains six space-separated integers.
The number 0 indicates an empty cell that you need to fill, while numbers 1 to 6 represent pre-filled cells.

\textbf{Output Format:}
Output the completed \sixsix{} grid, replacing the 0s with the correct numbers from 1 to 6, such that all the constraints are satisfied.

\textbf{Problem to Solve:}
\begin{quote}
\texttt{0 0 1 2 0 0 \\
0 0 0 0 0 0 \\
0 0 2 0 0 4 \\
0 0 0 6 0 5 \\
0 0 0 0 3 0 \\
4 0 0 0 0 6}
\end{quote}
Provide the completed \sixsix{} grid only
\end{minipage}
}
}

\section{Experiment Details}
\label{sec:experimentdetails}
The experiments were conducted using the following configurations:
\begin{enumerate}
    \item \textbf{Zero-shot Prompting:} For all experiments, we employed zero-shot prompting to evaluate the models' ability to solve 6x6 Sudoku puzzles without additional task-specific training or prior context.
    \item \textbf{Temperature and top-p:} All models were run with a fixed temperature of 0 to ensure deterministic outputs across multiple runs. Additionally we use top-p sampling value of 0.5 for all models.
    \item \textbf{Infrastructure:} Gemma-2-9B-Instruct was was run on GroqCloud. Llama-3.1-70B-Instruct, Llama-3.1-8B-Instruct, and Mistral-7B-Instruct-v0.3 models were accessed and run via the Hugging Face Chat API. o1-preview was used via OpenAI's user interface.
    \item \textbf{Accuracy Evaluation:} The number of pre-filled (givens) was also considered when calculating row, column, box and cell-wise accuracy.
\end{enumerate}

\section{Difficulty Distribution in 20-Puzzle Subset}
\label{sec:o1distribution}
For our detailed explanation analysis, we selected 20 puzzles from the larger dataset, ensuring representation across different difficulty levels.~\ref{fig:difficulty-dist-o1} shows the distribution of these puzzles across the Explainer Rating (ER) categories. This distribution was intentionally designed to be more balanced than the original dataset to ensure sufficient coverage of higher difficulty levels while maintaining representation of common puzzle types. The selection allows us to evaluate o1-preview's explanation capabilities across the full spectrum of puzzle complexity.

\section{o1-preview Performance Analysis by Difficulty Level and Evaluation Metric}
\label{sec:performanceo1difficulty}
\begin{figure}[!htbp]
    \centering
    \includegraphics[width=\linewidth]{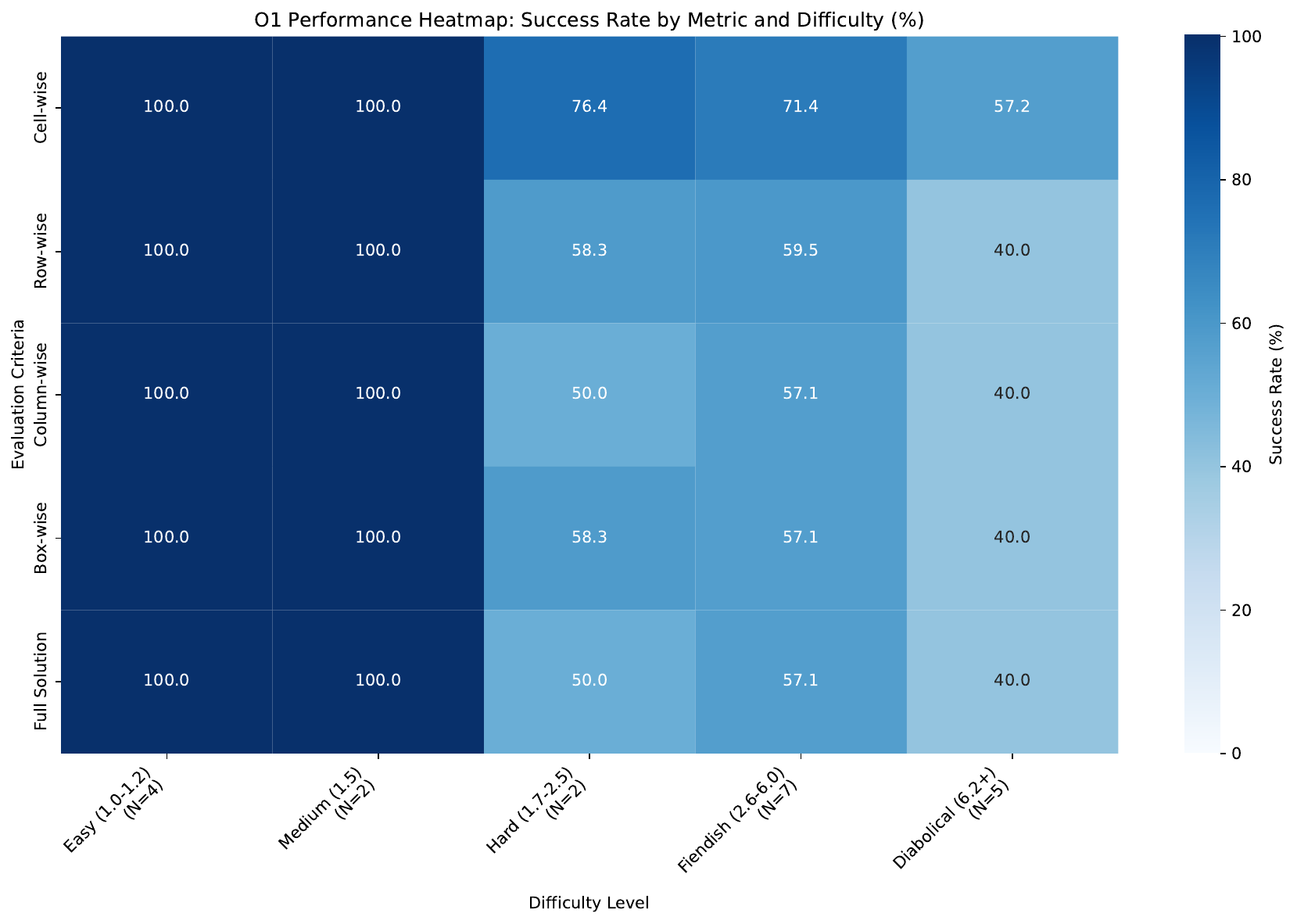}
    \caption{Heatmap visualization of o1-preview's performance across different difficulty levels and evaluation metrics for the 20-puzzle subset.}
    \label{fig:o1_performance_heatmap}
\end{figure}

Figure.~\ref{fig:o1_performance_heatmap} presents a detailed heatmap analysis of o1-preview's solving performance across the surveyed puzzles, revealing three key patterns:
\begin{itemize}
    \item \textbf{Perfect performance on simpler puzzles:} For Easy (ER 1.0-1.2) and Medium (ER 1.5) difficulty levels, o1-preview achieves 100\% accuracy across all metrics.
    \item \textbf{Progressive degradation with difficulty:} Performance declines notably for harder puzzles, with success rates dropping to 40\% for complete solutions in Diabolical puzzles (ER 6.2+).
    \item \textbf{Consistent metric hierarchy:} Cell-wise accuracy remains highest across all difficulty levels (57-100\%), while complete solution rates show the steepest decline (40-100\%), suggesting the model struggles more with global constraints than local patterns.
\end{itemize}

\section{Survey Participants and Expertise}
\label{sec:surveyexpertise}
Our Likert survey involved three participants from our institution: one faculty member and two graduate students, all of whom were connected to the research project. While their experience with Sudoku varied, all participants had a solid understanding of Sudoku rules and demonstrated experience in solving puzzles across different difficulty levels.

\end{document}

%% file: 1_intro.tex
\section{Introduction}
\label{sec:intro}
The recent success of Large Language Models (LLMs) in automating reasoning tasks using natural language highlights their potential in human-AI collaborative decision-support systems. 
However, their effectiveness in critical domains---such as safety-critical systems, legally sensitive decision-making, medical support systems, and business-critical processes---depends on their ability to provide step-by-step, consistent, and trustworthy explanations tailored to the prior knowledge and current capabilities of the human partner.
Solving complex, multi-step puzzles like Sudoku offers a meaningful example of such decision-making scenarios. 
These puzzles share several key challenges---including step-by-step reasoning, the need for tailored explanations, and the importance of maintaining consistency---but without the risks associated with making incorrect critical decisions. 
Sudoku thus provides a controlled environment to study the capabilities of LLMs in addressing these challenges while isolating them from high-stakes consequences.
This paper explores whether LLMs can effectively navigate these decision-making challenges by focusing on their performance in solving and explaining Sudoku problems. 

The ability of LLMs to reason and solve problems has become a key area of interest in the field of artificial intelligence~\cite{huang-chang-2023-towards,yin-etal-2024-reasoning}. Although LLMs have shown remarkable performance in tasks involving language generation, translation, and comprehension~\cite{NEURIPS2020_1457c0d6,10.5555/3600270.3602070}, their ability to handle more structured reasoning tasks is less understood~\cite{mirzadeh2024gsmsymbolicunderstandinglimitationsmathematical}. 
In the last year, several studies have attempted to test the abilities of LLMs to solve complex puzzles~\cite{chia-etal-2024-puzzlevqa,li-etal-2024-assessing-logical,giadikiaroglou2024puzzlesolvingusingreasoning}. While these models are increasingly capable of generating correct answers, the explanations they provide for their reasoning process often fall short~\cite{10.5555/3666122.3669397}. 
We argue that good reasoners should be able to clearly expose their deductive processes. Therefore, reasoning explanations should be faithful, provide enough support for individual decisions, and provide insight into the process they took to arrive at any given solution. 

While we can obtain correct answers and faithful reasoning steps for many puzzles with symbolic reasoners~\cite{Moura11}, it can be hard for humans to make sense of the explanations provided by these tools.  These explanations usually come in the form of fine-grained proofs that require a good grasp of proof terminology and are often overwhelmingly detailed.  They often refer to encodings of the problem that are cumbersome to relate to the original presentation of the puzzle.  To be useful in practical scenarios, explanations need to be easily understandable to humans~\cite{8631448}. Given that LLMs excel at producing human-like output~\cite{NEURIPS2022_b1efde53}, we are interested in studying the challenges and opportunities of LLMs for explaining puzzles solutions. 

\paragraph{Research Questions and Findings} 
We investigate the explanation capabilities of LLMs in the context of \sixsix{} Sudoku puzzles. Sudoku is a widely recognized logic-based puzzle~\cite{Rosenh11}, which provides an ideal benchmark to explore these facets of LLMs. Solving Sudoku requires deductive reasoning and rule-based elimination, skills that go beyond simple pattern recognition. We choose \sixsix{} Sudoku because, while not being overly difficult, it often requires a few nontrivial reasoning steps to be solved.  With our study, we aim to answer the following two research questions: (1) LLMs solve \sixsix{} Sudoku puzzles? (2) Can LLMs successfully explain the steps needed to arrive to a given solution? To answer these questions, we construct a dataset of 2,293 \sixsix{} Sudoku puzzles with their solutions, and show that open-source models can correctly solve less than 1\% of the puzzles, while the latest Open AI model can solve 65\% of them. Then, select a subset of 20 puzzles spanning different difficulty levels to evaluate explanation capabilities using a structured survey, and show that explanations fall short for the best performing models. 
\paragraph{Contributions} This study contributes to a deeper understanding of the (in)ability of LLMs to articulate the reasoning steps needed to explain a particular solution to multi-step reasoning tasks. We show that while these tools show promise (as evidenced by the increase performance for newer models), they still have significant challenges in exposing reasoning steps in a way that is both faithful and insightful. Finally, we outline a research agenda for combining LLMs with logic-based reasoning to address some of these challenges\footnote{All data and code will be released.}. 



%% file: 2_related.tex
\section{Related Work}
\label{sec:related}

The challenges that LLMs face in solving puzzles that require complex reasoning have been well-documented~\cite{giadikiaroglou-etal-2024-puzzle}, with most methods struggling to successfully produce consistent results. In response to these challenges, methods that combine symbolic reasoning and LLMs have recently received attention with varying success~\cite{mittal2024puzzlebench, Jiang2022DraftSA, Xin2024DeepSeekProverAT}.
PuzzleLM~\cite{mittal2024puzzlebench} first prompts an LLM to generate an instance-agnostic program capable of handling various problem classes, such as Sudoku, N-Queens, graph coloring, etc. The program then converts the problem instance into SMT constraints and passes them to an external solver like Z3. The solver processes these constraints and returns a solution, which the program then converts to the desired output format. ~\citet{Xin2024DeepSeekProverAT} introduces a method for generating large-scale formal proof data from informal math problems. It fine-tunes an LLM, translates informal problems into formal Lean 4 statements, filters low-quality formalizations using a chain-of-thought evaluation, and attempts to prove both the formal statement and its negation. This iterative process refines the model's performance by incorporating newly generated data to enhance subsequent rounds of statement generation. 

However, significant less attention has been given to the problem of whether LLMs can produce useful explanations (in form of reasoning steps) for the solutions that they provide. In this work, we study this question in the context of 6x6 sudoku and show that, while LLMs have improved their capacity of producing correct solutions, generating explanations continues to be challenging. 

%% file: 3_task.tex
\section{Task Formulation}
\label{sec:task}

This study assesses the problem-solving capabilities of LLMs using \sixsix{} Sudoku puzzles. We evaluate two key aspects: first, the ability of LLMs to generate correct solutions for puzzles of varying difficulty, and second, their capability to provide meaningful explanations for how these solutions are derived. These complementary tasks allow us to assess both the computational accuracy and reasoning transparency of LLMs in structured problem-solving scenarios. 

\paragraph{Dataset}
We generated 2,293 \sixsix{} Sudoku puzzles using the Z3 solver. Z3 decides the satisfiability of sentences of many decidable fragments of first-order logic. It combines the CDCL algorithm for propositional satisfiability with solvers for conjunctions of literals for \emph{background theories} like linear arithmetic.

\paragraph{Puzzle Generation Process} In our Sudoku generation process, each puzzle is modeled as a grid of cells, where every row, column, and $2\! \times\! 3$ box must contain the numbers 1 to 6 without repetition. 
 To do this, the generator initially evaluates random seeds, that is, partially filled grids with 11 givens, until Z3 confirms the existence of a solution (not necessarily unique).  Once a solved grid is found, givens are removed from it in random order until none can be removed lest the uniqueness of the solution should be lost.  Z3 is used to check that the solution remains unique.  Namely, if $x$ is the value in Row $y$ Column $z$ that is a candidate for removal, the constraint that asserts $rycz = x$ is replaced by $rycz \neq x$.  If the resulting constraints are unsatisfiable, the candidate can be dropped.  This results in a dataset whose every puzzle has a unique solution and is minimal, in the sense that removal of any clue introduces additional solutions.  
 
 The minimum number of givens for a 6x6 puzzle is 8~\cite{McGuire03042014}. While our generation process does not guarantee uniform sampling from all 28,200,960 valid 6x6 grids, we get essentially the same statistics for the number of givens when we start from a uniform random sample of grids. (We do not sample from the uniform distribution of grids in our generator because the approach is computationally infeasible for grids larger than 6x6.) Consequently, approximately 96\% of our resulting puzzles have between 9 and 11 givens. To assess the representativeness of our sample, we compare with publicly available benchmarks~\footnote{\url{https://www.menneske.no/sudoku/2x3/eng/showstartcount.html}}, which show the following distribution for a sample of 323,231 symmetrical 6x6 puzzles: 0.2\% puzzles with 8 givens; approx. 88.7\% with 10 or 12 givens, 11.1\% with 14 or more givens. Symmetrical puzzles have an even number of givens. Moreover, the puzzles in the Menneske data set with more than 8 givens, unlike ours, have redundant givens. Both differences contribute to the higher average number of givens when compared to our sample.


\paragraph{Assessing puzzle difficulty} To assess puzzle difficulty, we employ the widely-accepted Explainer Rating (ER) system, which rates puzzles based on the most complex solving technique required for their solution. As shown in Figure~\ref{fig:difficulty-dist}, the ER scale ranges from 1.0 (basic techniques) to above 6.2 (advanced techniques), with each solving technique assigned a specific rating value. Our puzzles fall into five distinct difficulty categories: Easy (ER 1.0-1.2), Medium (ER 1.5), Hard (ER 1.7-2.5), Fiendish(ER 2.6-6.0) and Diabolical (ER 6.2+). Further details can be found in App.~\ref{app:diff_rubric}.

\begin{figure}[t]
    \centering
    \includegraphics[width=\linewidth]{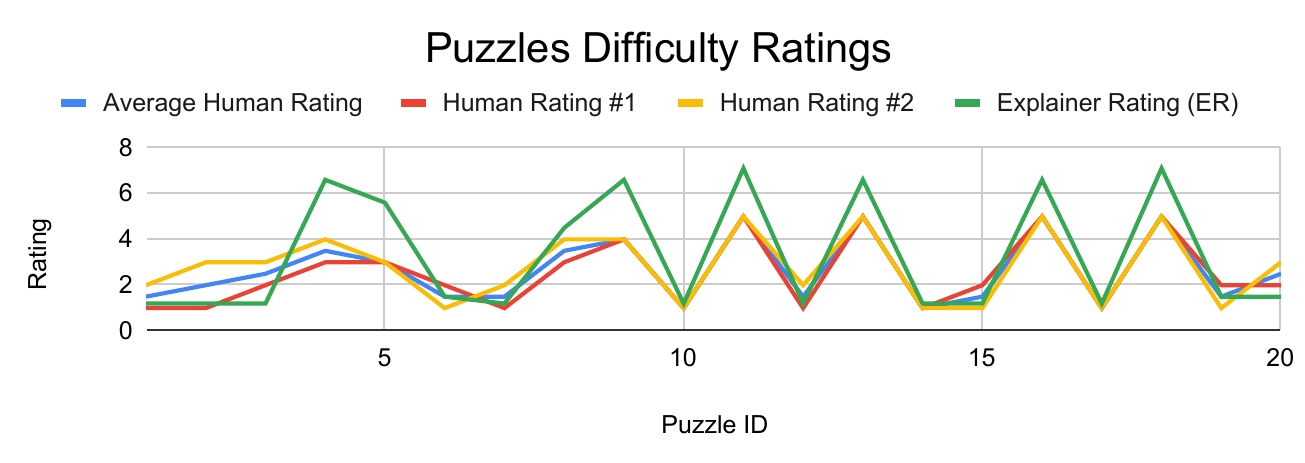}
    \caption{The difficulty rating for 2 Puzzles given by two human solvers and a Sudoku Explainer Program. A higher rating denotes higher difficulty.}
    \label{fig:HumanVsER}
\end{figure}

This distribution ensures our dataset spans a wide range of difficulty levels, with 93.5\% of puzzles being Easy or Medium difficulty, and 6.5\% requiring more advanced solving techniques. This variety allows us to evaluate both basic solving capabilities and more sophisticated reasoning strategies. To assess the reliability of this metric, 20 puzzles with varying numbers of back-jumps were solved and rated by two authors of the paper. The human solvers rated the puzzles on a scale from 1 to 5, with 1 denoting the easiest and 5 being the hardest among the set. Comparing the average rating of the two human solvers against the ER for the same puzzles resulted in a Spearman's Ranking Correlation Coefficient ($\rho$) of 0.86994. To put it in perspective, the two individual human rankings have a lower ($\rho$) of 0.8042. This ensured that the ER is reflective of the human perception of a puzzle's difficulty as seen in Fig. ~\ref{fig:HumanVsER}

%% file: 4_experiments.tex
\section{Experiments}
\label{sec:experiments}

\begin{table*}[ht]
\small
\centering
\begin{tabular}{l|c|c|c|c|c}
\toprule
& \multicolumn{5}{c}{Accuracy (\%)}\\\cline{2-6}
LLM                      & Cell-wise & Row-wise & Column-wise & Box-wise & Fully Correct Solutions \\ \midrule
Mistral-7B-Instruct-v0.3 & 21.96     & 0.59     & 0.17        & 0.17     & 0.00                    \\ 
Llama-3.1-8B-Instruct    & 39.65     & 2.22     & 2.01        & 1.36     & 0.00                    \\ 
Llama-3.1-70B-Instruct   & 45.93     & 4.59     & 6.29        & 2.70     & \textbf{0.04}           \\ 
Gemma-2-9B-Instruct      & \textbf{50.12} & \textbf{6.74} & \textbf{6.53} & \textbf{3.87} & \textbf{0.04} \\ \midrule
Mistral-7B-Instruct-v0.3 & 22.42     & 0.67     & 0.17        & 0.17     & 0.00                    \\ 
Llama-3.1-8B-Instruct    & 39.44     & 1.67     & 1.83        & 0.50     & 0.00                    \\ 
Llama-3.1-70B-Instruct   & 44.92     & 2.50     & 4.67        & 1.83     & 0.00                    \\ 
Gemma-2-9B-Instruct      & 49.47     & 5.33     & 6.00        & 2.33     & 0.00                    \\ 
OpenAI o1-preview        & \textbf{83.53} & \textbf{67.50} & \textbf{67.00} & \textbf{66.83} & \textbf{59.00} \\ \midrule
Humans                   & 100.00    & 100.00   & 100.00      & 100.00   & 100.00                  \\
\bottomrule
\end{tabular}%
\caption{Sudoku Solving Performance Across Models and Dataset Sizes. Upper section shows results for open-source models evaluated on the full dataset (2,293 puzzles). Lower section shows results for all models, including o1-preview, evaluated on a subset of 100 puzzles}
\label{tab:1}
\end{table*}




\paragraph{Can LLMs Solve Sudoku Puzzles?} We tested the performance of several state-of-the-art LLMs on our \sixsix{} Sudoku puzzles. The open-source models Llama-3.1-70B-Instruct and Llama-3.1-8B-Instruct~\cite{llama3group}, Gemma-2-9B-Instruct~\cite{Gemma2} and Mistral-7B-Instruct-v0.3~\cite{mistral7B} were evaluated on our full dataset of 2,293 puzzles. Due to the access limitations and rate restrictions of closed proprietary models, we evaluated OpenAI's o1-preview model on a randomly selected subset of 100 puzzles. 
Each model was tasked with generating solutions and the correctness of their outputs was evaluated at multiple levels: cell-wise, row-wise, column-wise, box-wise, and full accuracy. The prompt used can be found in  App.~\ref{sec:promptdetails}.  
 














Table~\ref{tab:1} shows the results split into two sections: performance on the full dataset (2,293 puzzles) for open-source models, and performance on the 100-puzzle subset for all models including o1-preview. 
Open source models only managed 0.04\% fully correct solutions, regardless of their size, showing that they struggle with integrating all constraints across the grid. For these models, column-wise and row-wise constraints were prioritized over box-wise constraints. Even then, they struggled enforcing constraints in general - exemplified by correctly placing 40-50\% of cells yet only satisfying 7\% of the row/column constraints. The results on the 100-puzzle subset revealed that OpenAI o1-preview considerably outperforms all open-source models, achieving 59\% accuracy for fully correct solutions compared to near-zero performance from other models. This significant performance gap suggests that proprietary models have made substantial progress in logical reasoning capabilities.

Performance analysis across difficulty levels reveals that o1-preview achieves 100\% accuracy for Easy (ER 1.0-1.2) and Medium (ER 1.5) puzzles, but shows significant degradation as complexity increases. For Diabolical puzzles (ER 6.2+), cell-wise accuracy drops to 57.2\% and complete solution accuracy to 40\%, suggesting the model struggles with integrating multiple constraints as puzzle complexity increases (detailed analysis in App.~\ref{sec:performanceo1difficulty})

\paragraph{Can LLMs Provide Good Explanations for Solutions?}
We conducted a detailed evaluation of o1-preview's explanation capabilities on a subset of 20 puzzles (difficulty distribution of these puzzles in App.~\ref{sec:o1distribution}, performance analysis in App.~\ref{sec:performanceo1difficulty}). Three expert evaluators (details on survey participators in App.~\ref{sec:surveyexpertise}) assessed the quality of o1's explanations using a structured survey, with each explanation being evaluated by two assessors.

The evaluation was conducted through surveying the correctness of the final solution, comments on the overall explanation, and a three-point Likert scale (Yes, Maybe, No) assessing three key dimensions: (1)
Justification (the explanation provides justification for the solution), (2) Clarity (the explanation is easy to follow and detailed), and (3) Educational Value (the explanation promotes understanding of Sudoku solving strategies). For each of the 20 puzzles, two evaluators independently rated o1's explanations across these three dimensions. This resulted in a total of 120 individual assessments (20 puzzles × 3 questions × 2 evaluators). We measured inter-annotator agreement using weighted Kappa and obtained 0.6 for educational value, 0.6 for justification, and 0.4 for clarity, which suggests moderate to good agreement.

While o1-preview achieved correct solutions in 65\% of cases, the evaluation of its explanations across 20 puzzles revealed significant limitations. Only 5\% of the responses indicate the explanation indeed justifies the solution with a "Yes" ratings (52.5\% "Maybe", 42.5\% "No"); while the model identified correct digit placements, it rarely justified why the placements were were chosen. The clarity dimension scored poorly, with 7.5\% "Yes" responses (32.5\% "Maybe", 60\% "No"), with reported issues regarding logical progression and inconsistent terminology. Educational value proved to be the weakest dimension, with just 2.5\% "Yes" responses (52.5\% "Maybe", 45\% "No"), as explanations failed to convey generalizable solving strategies. The qualitative feedback reinforced these findings – evaluators consistently noted that explanations lacked systematic progression, failed to justify moves, misused basic terminology, and poorly articulated the solution path. These results highlight a stark contrast between o1-preview's ability to solve puzzles and its capacity to explain its reasoning process effectively.

%% file: 5_conclusion.tex
\section{Conclusions and Future Work}
\label{sec:conclusion}
Our analysis of the performance of five LLMs has found that, with the exception of OpenAI o1-preview, they are not currently capable of solving \sixsix{} Sudoku puzzles.  OpenAI's LLM fares much better than the other LLMs as a solver, but does not provide explanations that may shed light on what deductions may be used to solve a puzzle. We argue that improving explanation generation for multi-step reasoning processes is a promising area of research, where the ability of LLMs to generate easy-to-understand language has great potential.


The ability to explain multi-step reasoning is of even greater value for decision processes that have unique sets of constraints and are much harder to solve and explain.  It is therefore justified to expect that LLMs may need the help of tools based on different forms of reasoning.  We are particularly interested in exploring approaches that combine LLMs with logic-based reasoning, as implemented in SMT solvers (e.g., Z3 or CVC5 \cite{Barbosa22}) and proof assistants like Lean \cite{Moura15} or Isabelle HOL \cite{Nipkow24}.  Explanations produced by these tools are usually reliable and detailed, but are notoriously hard for humans to parse and digest.  We expect LLMs to be able to interpret those explanations in ways accessible to the non-specialist and with levels of detail that match the needs of the human user. 

%% file: limitations.tex
\section{Limitations}
Although this study examines the capabilities of LLMs in solving and explaining \sixsix{} Sudoku puzzles, there are several limitations.

\begin{enumerate}

    \item \textbf{Dataset Scope:} Although we generated 2,293 puzzles using the Z3 solver, the subset used for human and o1-preview comparison was limited to 100 puzzles for solutions and 20 puzzles for explanations. This small experiment was enough to appreciate the inability of o1 to come up with satisfying explanations for its solutions. Given that our subsets contained puzzles of varying difficulty and that results were consistently deficient, we believe that we have enough evidence to assert that this is an area where cutting-edge LLMs struggle. However, a larger-scale study could strengthen these findings and provide a more comprehensive view of the models' overall capabilities and performance under different difficulty levels.

    \item \textbf{Puzzle Complexity:} We evaluated the models on \sixsix{} Sudoku puzzles, which, while requiring some non-trivial reasoning, are simpler than the more commonly studied $9\! \times\! 9$ puzzles. The limited complexity of the \sixsix{} grid may not fully capture the range of reasoning challenges that LLMs might encounter in more difficult puzzles, potentially underestimating the models’ limitations in handling more complex tasks.

    \item \textbf{Impact of Fine-Tuning:} Our study was focused on evaluating the capacity of LLMs to produce solutions and explanations out of the box, without any fine-tuning. Of particular interest was to study the disconnect between o1’s ability to solve puzzles vs. its inability to provide consistent, faithful explanations for its reasoning process. While fine-tuning models was out of scope for this work, it is possible that conclusions could differ if any training were to be performed. 

\end{enumerate}

%% file: acknowledgments.tex
\section*{Acknowledgments}

Ashutosh Trivedi holds the position of Royal Society Wolfson Visiting Fellow and acknowledges the support of the Wolfson Foundation and the Royal Society for this fellowship.